\documentclass[utf8]{frontiersSCNS} 

\usepackage{url,hyperref,lineno,microtype,subcaption}
\usepackage[onehalfspacing]{setspace}

\usepackage[OT1]{fontenc}
\usepackage{amsmath}
\usepackage[ruled,linesnumbered]{algorithm2e}

\DeclareMathOperator*{\argmax}{argmax} 

\def\keyFont{\fontsize{8}{11}\helveticabold }
\def\firstAuthorLast{Sample {et~al.}} 
\def\Authors{Xuetao Jiang\,$^{1}$, Meiyu Jiang\,$^{1}$,YuChun Gou$^{1}$, Qian Li,$^{2,*}$and Qingguo Zhou\,$^{1,*}$}
\def\Address{$^{1}$School of Information Science and Engineering, Lanzhou University, Lanzhou, China}
\def\Address{$^{2}$Shannxi Province Climate center, Xi'an, China}
\def\corrAuthor{Qian Li, Qingguo Zhou}

\def\corrEmail{liqian2011@163.com, zhouqg@lzu.edu.cn}

\begin{document}
\onecolumn
\firstpage{1}

\title[Forestry DT on ML]{Forestry digital twin with machine learning in Landsat 7 data}

\author[\firstAuthorLast ]{\Authors} 
\address{} 
\correspondance{} 

\extraAuth{}

\maketitle

\begin{abstract}
Modeling forests using historical data allows for more accurately evolution analysis, thus providing an important basis for other studies.
As a recognized and effective tool, remote sensing plays an important role in forestry analysis.
We can use it to derive information about the forest, including tree type, coverage and canopy density.
There are many forest time series modeling studies using statistic values, but few using remote sensing images. 
Image prediction digital twin is an implementation of digital twin, which aims to predict future images bases on historical data.
In this paper, we propose an LSTM-based digital twin approach for forest modeling, using Landsat 7 remote sensing image within 20 years.
The experimental results show that the prediction twin method in this paper can effectively predict the future images of study area.
    

\tiny
 \keyFont{ \section{Keywords:} Digital Twin, Remote Sensing, Machine Learning, Spatial Temporal Prediction} 
\end{abstract}

\section{Introduction}

\par
Forests play an important role in ecosystems. 
By analyzing changes in forest lands, researchers could find correlation in the area over time.
Time series data analysis of forest makes decision-making more efficient and reliable in ecology \cite{2013Woodland}. 
The development of computer technology brought many useful tools for researchers, by which they can model and analyze more accurately.
Digital modeling has become an attractive and practical topic among many fields.
The digital twin (DT) \cite{NEGRI2017939} represents the digital modeling of a real-world object. 
Theoretically, study of DT model is equivalent to the research on the corresponding actual object.
If DT technology is applied to forestry, the study area can be approximated by digital model.

\par
Because there is no forestry digital twin review article in our survey, we try to find some feasible approachs from a review of agricultural digital twin \cite{verdouw_digital_2021}.
Agricultural DT use images and point cloud data from agricultural machines and drones, therefore require powerful computer for data process \cite{rs13163169}.
Forest land analysis using remote sensing images which could be done in personal computer.
The agricultural DT requires the IoT devices on farms for real-time monitoring \cite{s22020498}, while forest land does not have such conditions, making it difficult to acquire additional data in forest.
Therefore, remote sensing images are indispensable data in forestry DT.
As a widely recognized tool for land surface characterization, remote sensing images are widely used in forestry analysis, including forest disturbance prediction \cite{Buma_2017}, tree species analysis \cite{tree_1593.1} and forest canopy \cite{JOSHI200684} analysis.
Yang et al \cite{yang_mapping_2021} constructed a forest map of the southern Great Plains using data from multiple satellites.
Healey et al \cite{healey_mapping_2018} used the random forest algorithm to measure forest variation.
Grabska et al \cite{grabska_evaluation_2020} used models such as XGB to predict forest species.
However, there is no study on spatial-temporal prediction of forest remote sensing images, i.e., predicting future forest image from historical image data.
Therefore, we propose an LSTM-based method to achieve forest image twin and forest prediction twin.

For the study area in this paper, we collect USGS Landsat 7 data from 2001 to 2021 at yearly intervals.
In one year, multi-view remote sensing images are processed to obtain a complete image of the study area by QGIS \cite{QGIS_software}. 
To reduce the depth of the machine learning model, we adopt the cropping algorithm to convert high-resolution images into small blocks.
Finally, the LSTM-based machine learning model is constructed to predict future image data from historical remote sensing image data. 
The process diagram of the project is as follows:

\begin{figure}[h!]
\begin{center}
\includegraphics[width=15cm]{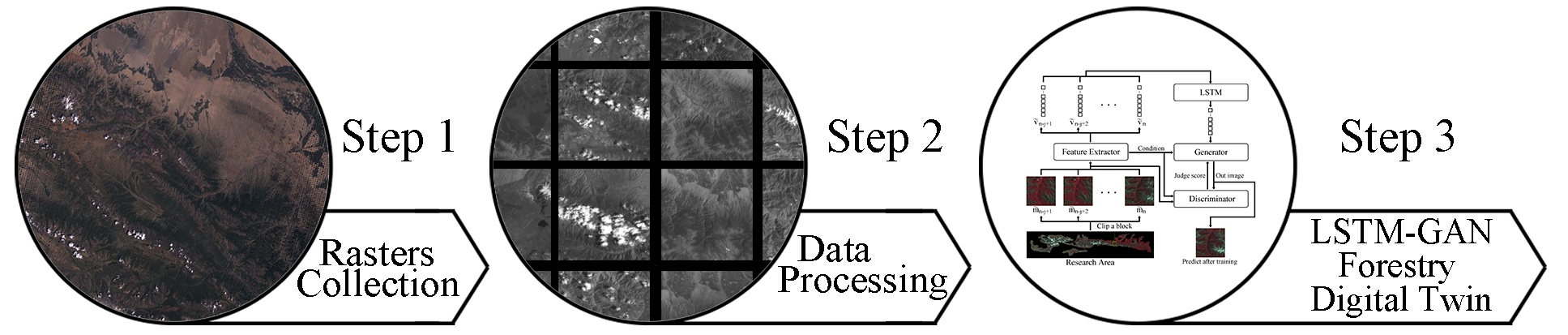}
\end{center}
\caption{Project work flow}\label{fig:1}
\end{figure}

This paper is organized as follows: Chapter 1 introduces the advantages of the forest DT and a method to implement it as presented.
Chapter 2 formulates the forestry image DT problem and presents the details of the LSTM-based model.
Chapter 3 analyzes the results to verify our method for forest DT.

\section{Method}
\subsection{Data Processing}

\begin{figure}[h!]
\begin{center}
\includegraphics[width=15cm]{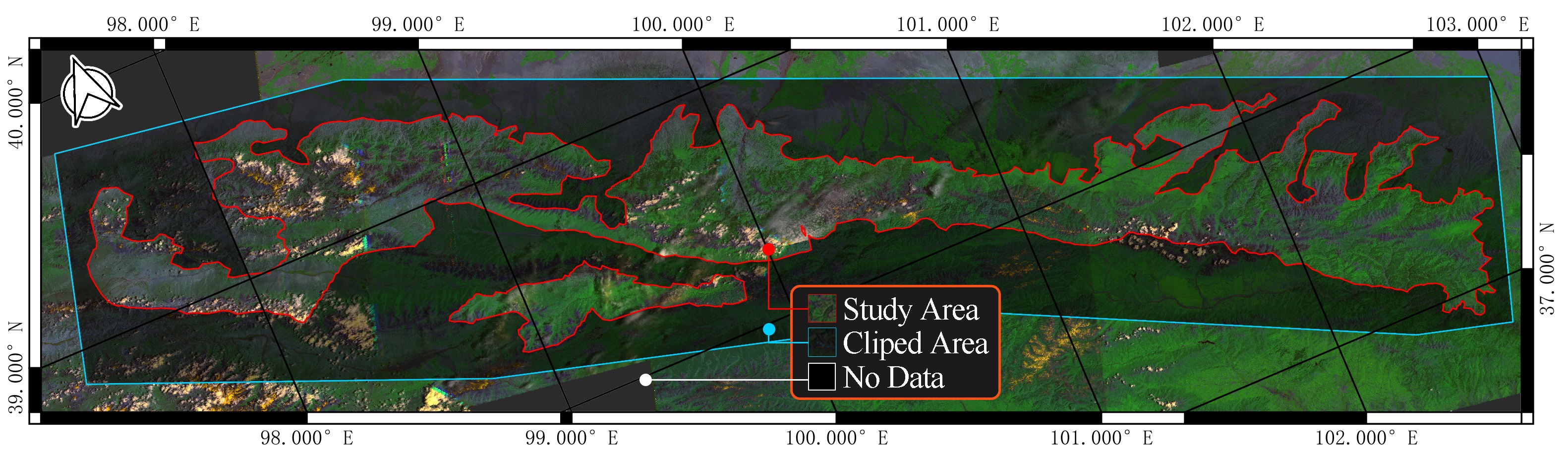}
\end{center}
\caption{Project work flow}\label{fig:2}
\end{figure}

The study area is a forest-stepp nature reserve in Gansu Province, China, as shown in Figure.2
First, we obtained the boundary of the area from the government website.
Then we collect remote sensing data on USGS Earth Explorer \cite{EarthExplorer} for free, covering 20 years from 2001 to 2021.
Due to the large area of reserve, 5 views of data are included every year.
And they are filled (gap filling), merged and clipped using QGIS to obtain the complete study area by GDAL library \cite{GDAL} and GRASS7 library \cite{GRASS_GIS_software}.
Finally 21 large remote sensing images were obtained.

However, modeling using the above images may suffer from insufficient data, oversized models and low accuracy.
We applied a cropping algorithm to slice a large image into smaller ones.
To ensure that the machine learning model can use all the pixels in the study area, the cliped area is an extension of the study area as shown in Figure 2.
Since the cropping algorithm in this paper uses tricks such as sliding window, variable window size and joint domain detection, describing these algorithms may take up a lot of space.
We represent the key steps of the algorithm in Figure. 3.

\begin{figure}[h!]
\begin{center}
\includegraphics[width=15cm]{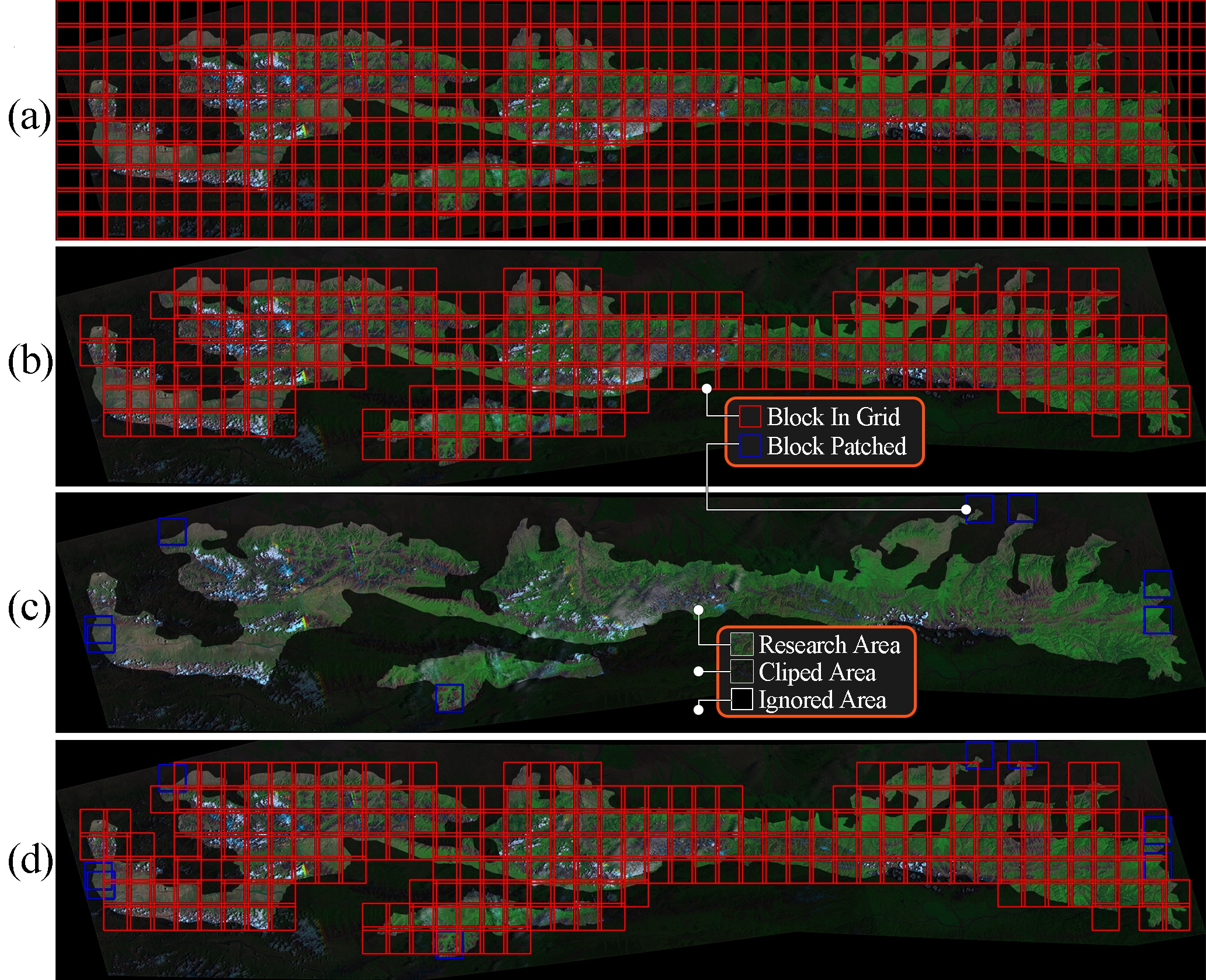}
\end{center}
\caption{Cropping remote sensing images}\label{fig:3}
\end{figure}

As shown in Figure 3 (a), we first traverse the entire region based on the cut size and move step.
It can be seen that the adjacent data blocks have overlaps, which are designed for smoothing in the final prediction.
After the traversal, we discard the data blocks that do not contain any pixels of the study region as shown in Figure 3 (b).
For uncovered area of study regions, we apply a gradient ascent strategy to find the data blocks as few steps as possible, and the results are shown in Figure 3 (c).
As shown in Figure 3 (d), the output of the algorithm is the coordinates used for Cropping.
Finally, we cut the images of each band to generate a dataset, which will be for subsequent model training.
There is a total of 42,840 image in the dataset, which comes from 8 bands.
Each one is a grayscale image of shape 128x128.

\subsection{Forest Image DT formulation}
In this section, we provide a formulaic description of forest image DT, which gives the theoretical basis for the model design.
Forest image DT uses historical remote sensing data for modeling, which aims to predict later frames by former frames.
Reducing the variability of data can improve model's robustness, we regularize each block separately.
For each block, we calcuate the mean and variance of the images over 20 years, and apply max-min regularization on them.
The regularized input image can be represented as a square matrix $m \in \mathbf{R}^{n}$.
Suppose the total frames of observable data is $n$, where $j$ consecutive observations are denoted as $\tilde{m}_{n-j+1} \cdots \tilde{m}_{n}$.
The n+1st data $m_{n+1}$ is unobservable and the corresponding maximum estimate is denoted as $\hat{m}_{n+1}$.
Then the prediction of the next data from j consecutive observations can be expressed as:

\begin{equation}
\hat{m}_{n+1} = \argmax_{m_{n+1}} P(m_{n+1}|\tilde{m}_{n-j+1} \cdots \tilde{m}_{n})
\end{equation}

\subsection{Forest Image DT Model}

Our model consists of three components: a feature extraction network, a long-short memory network (LSTM) \cite{LSTM}, and a generative network.
The feature extraction network uses a convolutional structure to downscale the remote sensing image data.
The LSTM uses several downscaled neighboring frames as input, then outputs the feature vectors of the predicted frames.
The generative network uses the features obtained from the LSTM network to generate the predicted frames.

\subsubsection{LSTM Model and Feature Extraction Model}

For each block, the order of observable historical images $\tilde{m}$ is 16,384.
As a picture it is low dimensional, but for time series it has a high dimensionality.
We therefore used a two-layer convolutional network $f$ as feature extraction network to reduce the dimensionality of each image data.
In our experiments, without feature extraction network, the training of the other the models become unstable.
If the images are input directly to the LSTM network for temporal feature extraction, the training loss converges very slowly.

\begin{equation}
\tilde{v}_i = f(\tilde{m}_i), i \in {n-j+1 \cdots n}
\end{equation}

Through the feature extraction network $f$, the dimensionality of feature vector $v$ is much smaller than image $\tilde{m}$.
In the next step, we use the LSTM model to extract the temporal features from the reduced vectors.
LSTM \cite{LSTM} is a special recurrent neural network  which is good at learning the dependencies from long sequences.
The key structure of the LSTM is the cell state, which control the forgetting or retention of the newly added state.
The LSTM outputs the final state by accepting a number of consecutive historical data h.

\begin{equation}
z = lstm(\tilde{v}_{n-j+1},\cdots,\tilde{v}_{n})
\end{equation}

The final state $z$ obtained by LSTM through vectors $\tilde{v}_{n-j+1} \cdots \tilde{v}_{n}$ is the temporal feature vector for GAN model.

\subsubsection{Generator Model}
Generative adversarial network (GAN) \cite{NIPS2014_5ca3e9b1} could generate new data similar to the original data, which is consists of a generator and a discriminator.
However, GAN aims to generate similar images on the dataset.
Frankly, the new images generated by GAN contain the features of several images in the dataset, but may not be similar to a particular image.
In our DT model, we want the predicted image to be as similar as possible to the actual image.
Therefore, we only used the generator in GAN and optimized it together with the other models.
Since the predicted images are divided into many blocks, we use the design of conditional GAN [CITE](CGAN) in the model.
We use the coordinates of the data as the conditions, e.g., 0111 denotes the seventh block in grid.
By adding the additional condition, our method can model many blocks at same time.

\subsubsection{Model Design}

The model in this paper consists of the above three parts, and its structure is shown in Figure.4.

\begin{figure}[hbt!]
\begin{center}
\includegraphics[width=15cm]{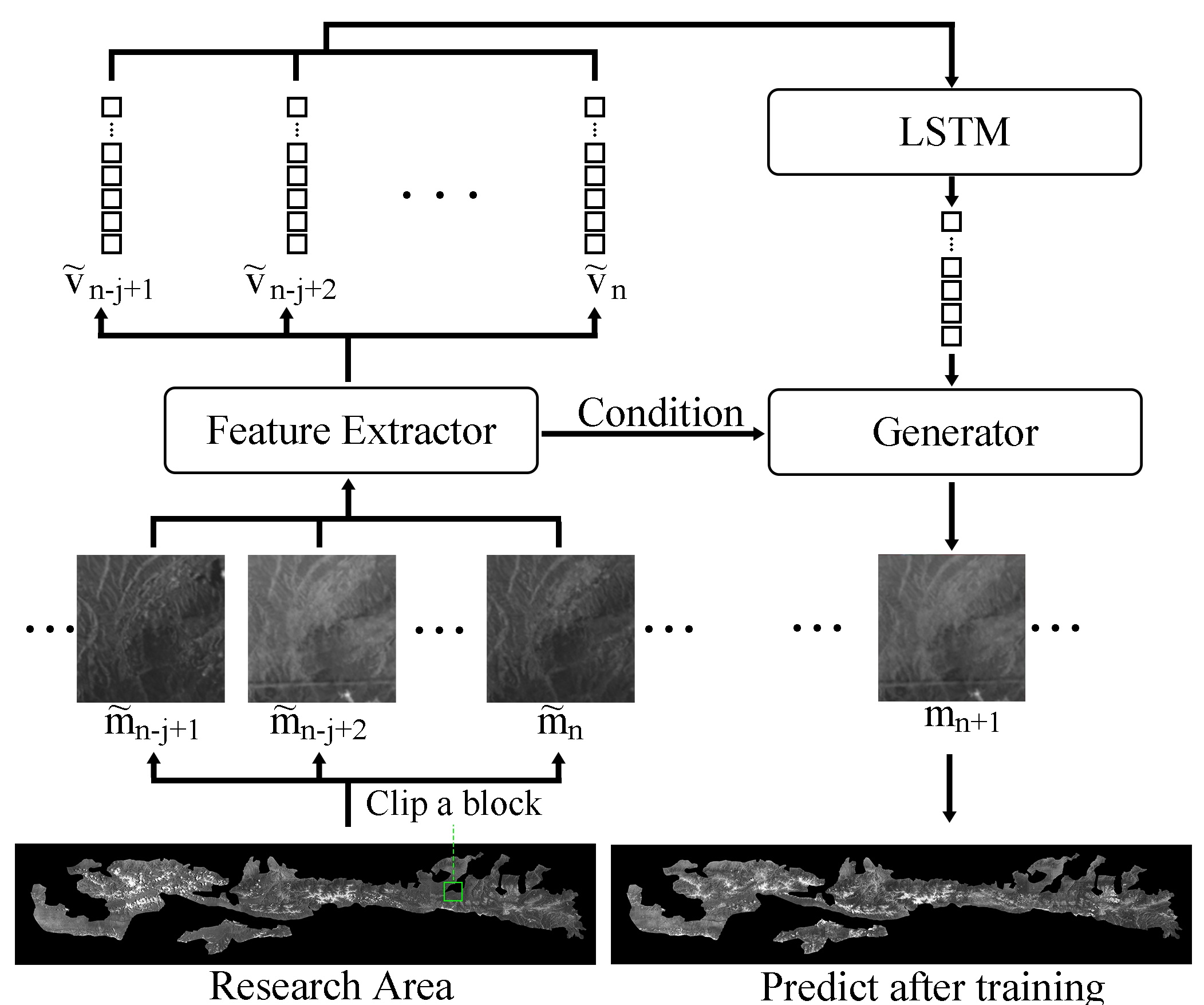}
\end{center}
\caption{Data flow of the model}\label{fig:4}
\end{figure}

Since we use a generator to rebuild future image, $m_{n+1}$ can be replaced by $G(z,c)$ in Equation 1.
Next, the feature extraction network extracts downscaled $\tilde{m}_{n-j+1} \cdots \tilde{m}_{n}$ into $\tilde{v}_{n-j+1} \cdots \tilde{v}_{n}$.
Feature vectors $\tilde{v}_{n-j+1} \cdots \tilde{v}_{n}$ is associated with the generating condition $z$ using LSTM model.
The final result is an approximation of the actual image, achieving the goal of DT model.
Finally, equation 1 can be expressed as:

\begin{align}
\hat{m}_{n+1} &= \argmax_z p(G(z)| \tilde{m}_{n-j+1} \cdots \tilde{m}_{n}) \nonumber\\
  &\approx \argmax_z p(G(z)| \tilde{v}_{n-j+1} \cdots \tilde{v}_{n}) \\ 
  &\approx G(lstm(\tilde{v}_{n-j+1},\cdots ,\tilde{v}_{n})) \nonumber 
\end{align}

\subsection{Model Training}

Many researchs also use generator as part of the model, which has two different training methods.
The first way trains the GAN alone, then combine pre-trained generator and the rest model for training.
The second way trains the generator with the other parts simultaneously.
In the beginning, we trained CGAN model to obtain the generator, and combine it with the other part for training.
However, during the training of CGAN, the input of generator is the noise within Gaussian distribution.
We added a parametric Gaussian distribution layer to the end of LSTM, but the training of the total model become unstable and collapse frequently.
Next we tried the second method and found it to be more robust.
Therefore, we use a uniform training approach, which is described as follows.

\begin{algorithm}
\caption{Model training}\label{alg:1}
\KwData{feature extractor part $f$, LSTM part $lstm$, generator part $G$, training epochs $itr$, data set $ds$}
\KwResult{Trained model $Model(f,lstm,G)$}
\For{$i\; \mathrm{in} \; itr$}{
    \For{$\{[\tilde{m}_1, \tilde{m}_2, \cdots, \tilde{m}_j,], m_{j+1}, c\}\; i\; \mathrm{in} \;  ds $}
    {
        $[\tilde{v}_1, \tilde{v}_2, \cdots, \tilde{v}_j,] \gets f([\tilde{m}_1, \tilde{m}_2, \cdots, \tilde{m}_j,])$ \;
        $z \gets lstm([\tilde{v}_1, \tilde{v}_2, \cdots, \tilde{v}_j,])$ \;
        loss = MSE(G(z,c),$m_{j+1}$) \;
        G.back(loss)\;
        lstm.back(loss)\;
        f.back(loss)\;
    }
}
\Return{Model(f,lstm,G)}
\end{algorithm}

Algorithm 1 takes the feature extract network $f$, LSTM network $lstm$, generator network $G$, training epochs $itr$ and data set $ds$ as inputs. 
And it outputs the trained model in the end.
Line 1 to line 10 indicates training the model $itr$ times.
Line 3 indicates that the input matrices are compressed into feature vectors by feature extract network.
Line 4 indicates that the feature vector will be transformed into a temporal feature sequence through the LSTM.
Line 6 to line 9 indicates that the three networks are trained sequentially by back-propagating the loss.

\section{Result and Analyze}
Analysis on remote sensing images require many different band combinations.
Here we use the B432 combination as an example.
In this paper, data from 2001 to 2020 were used for training and data from 2021 were used for prediction.
The result comparison is shown below.

\begin{figure}[hbt!]
\begin{center}
\includegraphics[width=15cm]{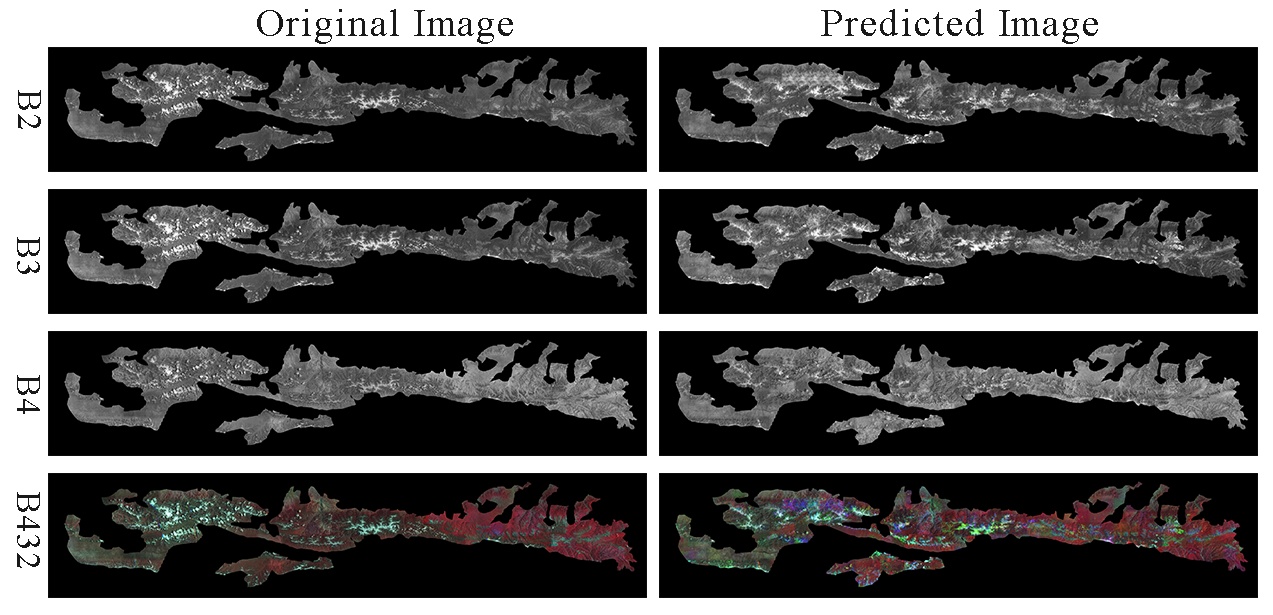}
\end{center}
\caption{Model prediction for B2, B3 and B4}\label{fig:5}
\end{figure}

From the figure we can see that the generated image is very similar to the predicted image.
To further analyze their similarity, we evaluated them using the NRMSE score.

\begin{equation}
NRMSE = \sqrt\frac{\sum_{i=1}^{n}   \left(y_{i} - \hat{y}\right)^{2}}   {n\sigma^2 }
\end{equation}

where $y_i$ denotes the true value, $\bar{y}$ denotes the mean of the true value, $\hat{y_i}$ denotes the predicted value, and $\sigma$ denotes the standard deviation.
We calculate NRMSE score for each band as shown in Figure. 6.

\begin{figure}[hbt!]
\begin{center}
\includegraphics[width=15cm]{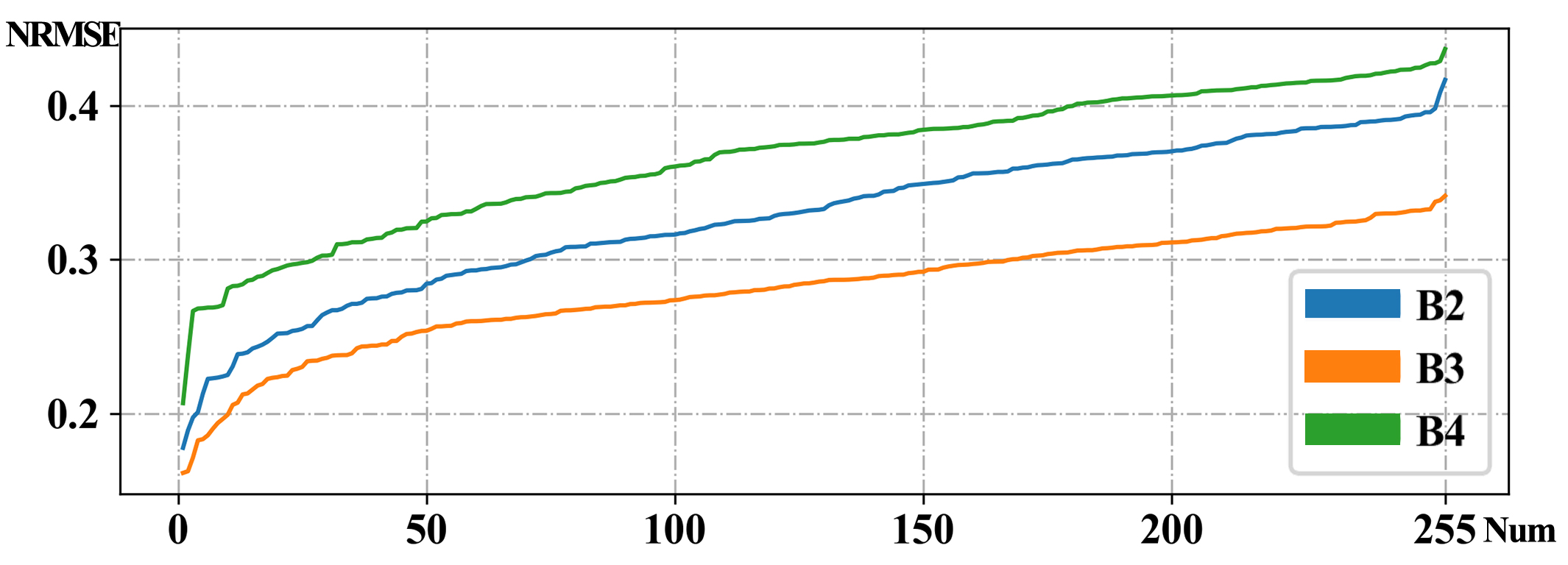}
\end{center}
\caption{NRMSE score of each band}\label{fig:6}
\end{figure}

It can be seen that the three curves have the same trend, with B3 $>$ B2 $>$ B4 in total area.
To further analyze the regions with different scores, we quadrupled the scores for each band, and plotted their original and predicted maps in Figure 7.

\begin{figure}[hbt!]
\begin{center}
\includegraphics[width=15cm]{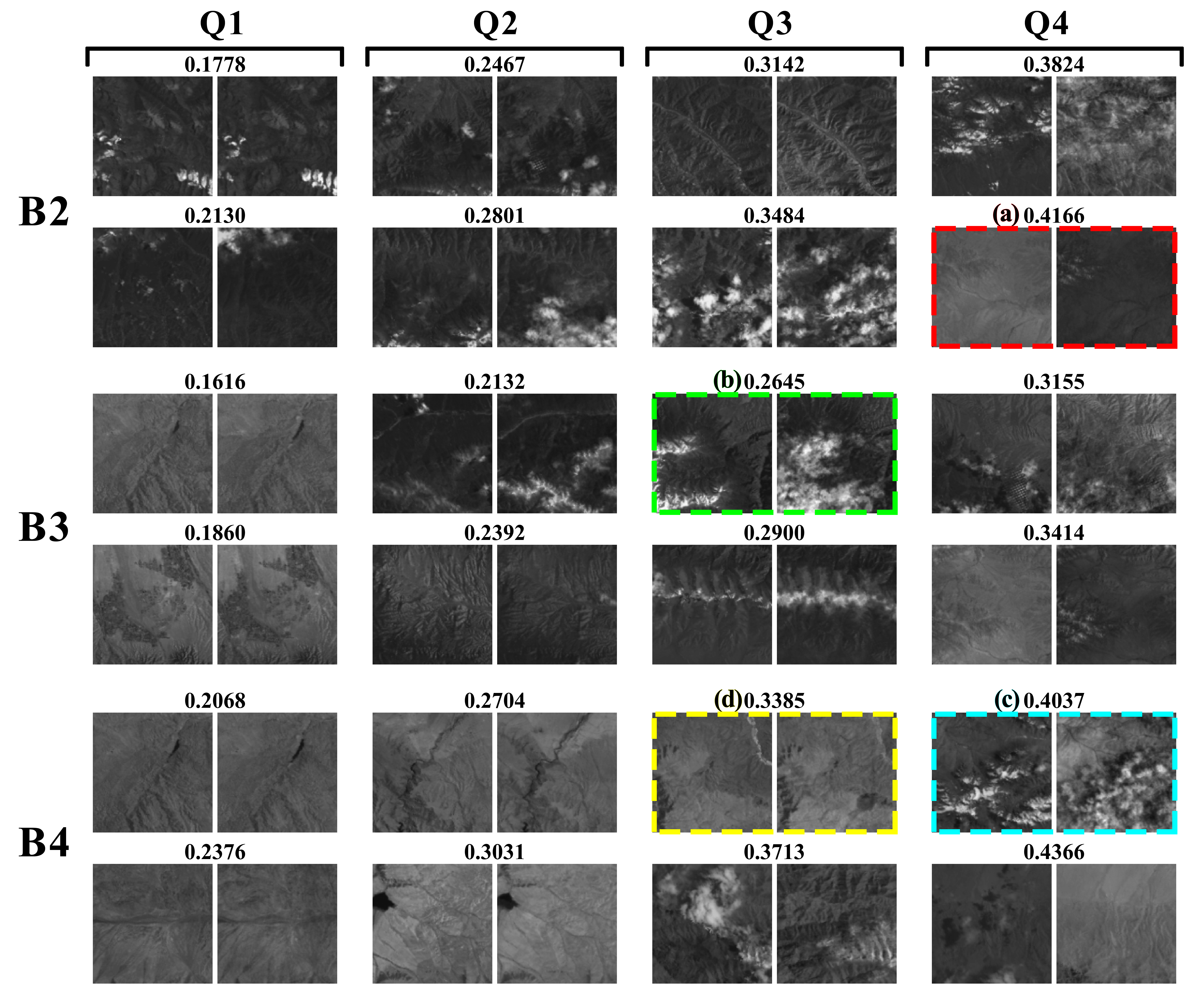}
\end{center}
\caption{NRMSE scores and the prediction}\label{fig:7}
\end{figure}

In Figure 7, there are eight pairs of images for each band, equally divided into four intervals.
For each pari of images, the left one is the original image and the right one is the predicted image.
Obviously, a low score indicates a high similarity between the two images.
The predicted and true values are very close in columns Q1 and Q2.
The predictions in column Q3 and column Q4 are biased, but still have some common region with true value.
We selected four representative dissimilar conditions, which are indicated by colored dashed boxes in Figure 7.
(a) shows that the prediction and the original image are not in one color interval. For this block, it have different color distribution from 2018 to 2020. This may lead to bias in the prediction interval of the model.
(b) shows that the prediction has additional cloud. 
For remote sensing data with long time spans over large areas, it is almost impossible to guarantee that every piece of data is cloud-free.
If meteorological data for the region are not available, the cloud coverage model in the region cannot be built.
(c) is the simultaneous occurrence of case (a) and case (b).
(d) shows that the changes in the region are not reflected in the prediction.
For this region, the changes that occur in 2021 did not occur in previous years.
So this is an unavoidable bias.

Therefore, the inconsistency between the predicted and original images may be influenced by the inconsistency of data.
In the absence of meteorological data such as local light radiation data, it is difficult to optimize the model to get better prediction.

\section{Conclusion}

In this paper, a digital twin method for forestry image prediction is designed to predict future remote sensing data by using historical LandSat 7 remote sensing data.
A forestry image chunking method was designed to slice the large scale remote sensing images into smaller chunks according to the study area, so that the model can model the large scale forestry images.
An LSTM-based model was designed to predict future remotely sensed images using remote sensing image time series for training.
From the prediction results, it can be seen that the method can predict the development of forestry images to a certain extent and works well as a forestry prediction twin.

\section*{Conflict of Interest Statement}

The authors declare that the research was conducted in the absence of any commercial or financial relationships that could be construed as a potential conflict of interest.

\section*{Author Contributions}

The Author Contributions section is mandatory for all articles, including articles by sole authors. If an appropriate statement is not provided on submission, a standard one will be inserted during the production process. The Author Contributions statement must describe the contributions of individual authors referred to by their initials and, in doing so, all authors agree to be accountable for the content of the work. Please see  \href{http://home.frontiersin.org/about/author-guidelines#AuthorandContributors}{here} for full authorship criteria.

\section*{Funding}
This work has no funding.

\section*{Acknowledgments}
This work was done by members of the dslab lab in the School of Information Science and Engineering at Lanzhou University.



\bibliographystyle{frontiersinSCNS_ENG_HUMS} 
\bibliography{ref}


\section*{Figure captions}





\end{document}